\documentclass[11pt,a4paper]{article}

\usepackage[hyperref]{emnlp2020}
\usepackage{times}
\usepackage{latexsym}

\usepackage{microtype}

\usepackage{amsmath}
\usepackage{amssymb}

\usepackage{caption}
\usepackage{subcaption}
\usepackage{comment}

\usepackage{pifont}

\usepackage{paralist}
\usepackage{mathtools}
\usepackage{multirow}
\usepackage[capitalise]{cleveref}
\usepackage{booktabs}
\usepackage{svg}

\DeclareMathOperator*{\argmax}{arg\,max}

\newcommand{\ie}{i.e.,}

\newcommand{\localmethod}{\textsc{TowerBuilder}}
\newcommand{\globalmethod}{\textsc{SkylineBuilder(-RL)}}
\newcommand{\globalmethodRL}{\textsc{SkylineBuilder}}

\fi

\usepackage{glossaries}
\glsdisablehyper

\newacronym{ODQA}{ODQA}{Open-Domain Question Answering}
\newacronym{MRS}{MRS}{Machine Reading at Scale}
\newacronym{ACT}{ACT}{Adaptive Computation Time}
\newacronym{UT}{UT}{Universal Transformers}

\aclfinalcopy 
\def\aclpaperid{1837} 

\title{{D}on't {R}ead {T}oo {M}uch {I}nto {I}t:\\{A}daptive {C}omputation for {O}pen-{D}omain {Q}uestion {A}nswering}

\author{
  Yuxiang Wu \quad Sebastian Riedel \quad Pasquale Minervini \quad Pontus Stenetorp\\
  University College London \\
  { \normalsize \tt 
  \{yuxiang.wu,s.riedel,p.minervini,p.stenetorp\}@cs.ucl.ac.uk}
}

\date{}

\begin{document}
\maketitle
\begin{abstract}
Most approaches to \acrlong{ODQA} consist of a light-weight retriever that selects a set of candidate passages,
and a computationally expensive reader that examines the passages to identify the correct answer.
Previous works have shown that as the number of retrieved passages increases, so does the performance of the reader.
However, they assume all retrieved passages are of equal importance and allocate the same amount of computation to them, leading to a substantial increase in computational cost.
To reduce this cost, we propose the use of \emph{adaptive computation} to control the computational budget allocated for the passages to be read. 
We first introduce 
a technique operating on individual passages in isolation which relies on anytime prediction and a per-layer estimation of an early exit probability. 
We then introduce \globalmethodRL{}, an approach 
for dynamically deciding on which passage to allocate computation at each step, based on a resource allocation policy trained via reinforcement learning.
Our results on SQuAD-Open show that adaptive computation with global prioritisation improves over several strong static and adaptive methods, leading to a 4.3x reduction in computation while retaining 95\% performance of the full model.  
\end{abstract}

\section{Introduction}
\acrfull{ODQA} requires a system to answer questions using a large collection of documents as the information source. 
In contrast to context-based machine comprehension, where models are to extract answers from single paragraphs or documents, it poses a fundamental technical challenge in \emph{machine reading at scale}~\citep{DBLP:conf/acl/ChenFWB17} .
\begin{figure*}[t]
\begin{center}
\includegraphics[height=5.3cm]{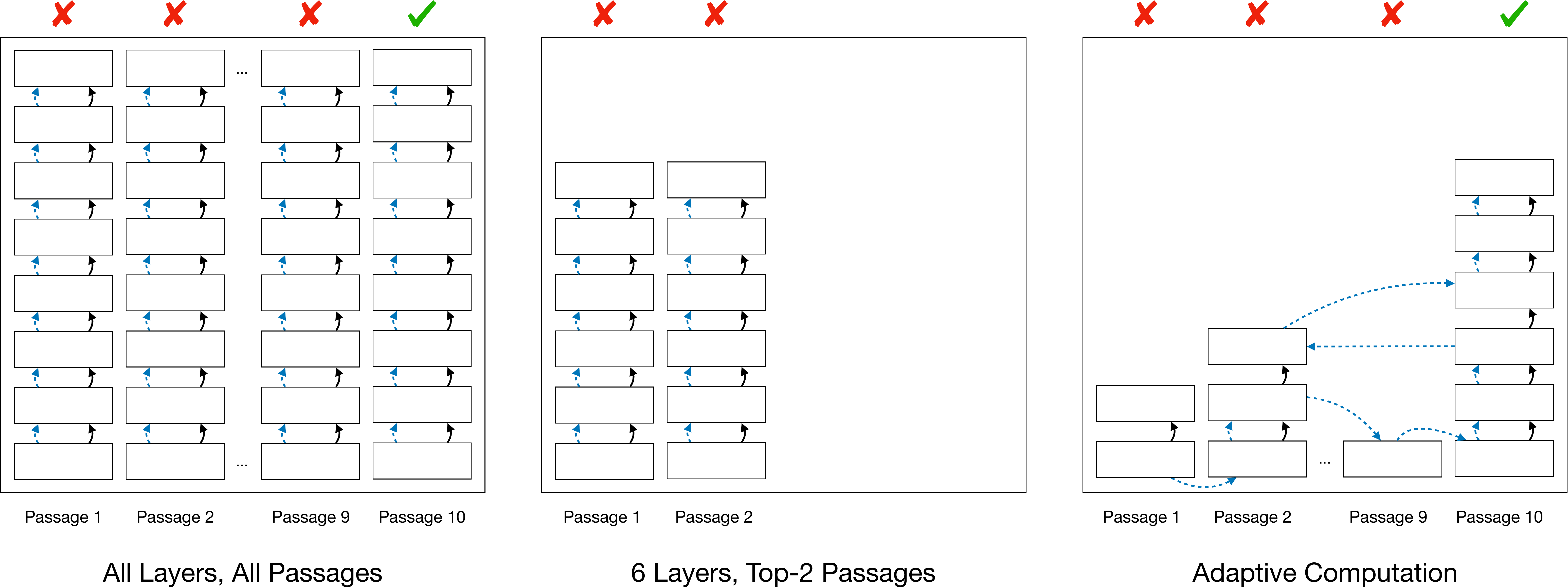}
\end{center}
\caption{Static and adaptive computation for Open-Domain QA. Each block represents one layer of transformer computation on a passage. The solid arrows show how activations flow, and the dashed arrows indicate the order of computation. Only passage 10 contains the actual answer. Using all layers on all passages can find the answer, while processing only the top 2 retrieved passages with 6 layers is unable to find it. 
Adaptive computation can find the right passage, and allocates most computation budget to reading it.}
\label{fig:splash}
\end{figure*}

Most \acrshort{ODQA} systems consist of two-stage pipelines, where
\begin{inparaenum}[\itshape 1\upshape)]
\item a context retriever such as BM25~\citep{DBLP:journals/jd/Robertson04} or DPR~\citep{DBLP:journals/corr/abs-2004-04906} first selects a small subset of passages that are likely to contain the answer to the question, and
\item a machine reader such as BERT~\cite{DBLP:conf/naacl/DevlinCLT19} then examines the retrieved contexts to extract the answer.
\end{inparaenum}
This two-stage process leads to a computational trade-off that is indicated in \cref{fig:splash}. We can run computationally expensive deep networks on a large number of passages to increase the probability that we find the right answer (``All Layers, All Passages''), or cut the number of passages and layers to reduce the computational footprint at the possible cost of missing an answer (``6 Layers, Top-2 Passages'').  
We hypothesise that a better accuracy-efficiency trade-off can be found if the computational budget is not allocated statically, but based on the complexity of each passage, see ``Adaptive Computation'' in \cref{fig:splash}. If a passage is likely to contain the answer, allocate more computation. If it isn't, allocate less. The idea of conditioning neural network computation based on inputs has been pursued in previous work on \emph{Adaptive Computation}~\citep{DBLP:journals/corr/BengioBPP15,DBLP:journals/corr/Graves16,DBLP:conf/iclr/ElbayadGGA20}, 
however how to apply this idea to \acrshort{ODQA} is still an open research question.

In this work, we introduce two adaptive computation methods for \acrshort{ODQA}: \localmethod{} and \globalmethodRL{}.
\localmethod{} builds a \emph{tower}, a composition of transformer layers on a single passage, until an early stopping condition is met---we find that this method already helps reducing the computational cost required for reading the retrieved passages. 
Then, for coordinating the construction of multiple towers in parallel, we 
introduce a global method, \globalmethodRL{}, that incrementally builds multiple towers one layer at a time and learns a policy to decide which tower to extend one more layer next. 
Rather than building single transformer towers in isolation, it constructs a \emph{skyline} of towers with different heights, based on which passages seem most promising to process further.
Our experiments on the SQuAD-Open 
dataset 
show that our methods are very effective at reducing the computational footprint of \acrshort{ODQA} models. 
In particular, we find that \globalmethodRL{} 
retains 95\% of the accuracy of a 24-layer model using only 5.6 layers on average.
In comparison, an adaptation of the method proposed by \citet{DBLP:conf/acl/SchwartzSSDS20} requires 9 layers for achieving the same results.
Improvements are even more substantial for smaller number of layers---for example, with an average of 3 layers \globalmethodRL{} reaches 89\% of the full performance, whereas the approach of \citet{DBLP:conf/acl/SchwartzSSDS20} yields 57\% and a model trained to use exactly 3 layers reaches 65\%. 
Finally, \globalmethodRL{} retains nearly the same accuracy at full layer count. 
To summarise, we make the following contributions: 
\begin{inparaenum}[\itshape 1\upshape)]
\item{
    we are the first to explore adaptive computation for \acrshort{ODQA} by proposing two models: $\localmethod$ and $\globalmethodRL$; 
}
\item{
    we experimentally show that both methods can be used for adaptively allocating computational resources so to retain the predictive accuracy with a significantly lower cost, and that coordinating the building of multiple towers via a learned policy yields more accurate results;
}
\item{
    when compared to their non-adaptive counterparts, our proposed methods can reduce the amount of computation by as much as 4.3 times.
}
\end{inparaenum}

\section{Background}

\newcommand{\layer}{\ensuremath{\mathrm{TransformerLayer}}}
\newcommand{\outputlayer}{\ensuremath{\mathrm{OutputLayer}}}
\newcommand{\logitslayer}{\ensuremath{\mathrm{Logits}}}
\newcommand{\softmax}{\ensuremath{\mathrm{Softmax}}}
\newcommand{\sigmoid}{\ensuremath{\mathrm{Sigmoid}}}
\newcommand{\mlp}{\ensuremath{\mathrm{MLP}}}

\newcommand{\problayer}{\ensuremath{\mathrm{P}_{\text{out}}}}
\newcommand{\logits}{\ensuremath{\mathbf{s}}}
\newcommand{\x}{\ensuremath{\mathbf{x}}}
\newcommand{\threshold}{\ensuremath{\tau}}
\newcommand{\noanswer}{\ensuremath{\mathrm{NoAnswer}}}
\newcommand{\hasanswer}{\ensuremath{\mathrm{HasAnswer}}}

\newcommand{\hidden}{\ensuremath{\mathbf{a}}}
\newcommand{\hiddent}{\ensuremath{\mathbf{h}}}

\newcommand{\result}{\ensuremath{\mathbf{y}}}
\newcommand{\query}{\ensuremath{\mathbf{q}}}
\newcommand{\corpus}{\ensuremath{C}}
\newcommand{\docset}{\ensuremath{D}}
\newcommand{\reader}{\ensuremath{\mathrm{Reader}}}
\newcommand{\passagereader}{\ensuremath{\mathrm{PReader}}}
\newcommand{\aggregate}{\ensuremath{\mathrm{Agg}}}
\newcommand{\CLS}{\ensuremath{\mathrm{CLS}}}
\newcommand{\currepr}{\ensuremath{A}}
\newcommand{\curheight}{\ensuremath{H}}
\newcommand{\state}{\ensuremath{S}}
\newcommand{\height}{\ensuremath{h}}
\newcommand{\buildupphase}{\ensuremath{\mathrm{BuildUp}}}
\newcommand{\outputphase}{\ensuremath{\mathrm{Output}}}

\newcommand{\numoutputs}{\ensuremath{m}}
\newcommand{\priority}{\ensuremath{p}}
\newcommand{\heightembed}{\ensuremath{\mathrm{HeightEmb}}}
\newcommand{\rankembed}{\ensuremath{\mathrm{IndexEmb}}}
\newcommand{\action}{\ensuremath{\mathbf{i}}}
\newcommand{\prioritymodel}{\ensuremath{\mathrm{PriorityModel}}}
\newcommand{\alllayer}{\ensuremath{\mathrm{AnyLayer}}}
\newcommand{\lastlayer}{\ensuremath{\mathrm{LastLayer}}}

\newcommand{\devzero}{\ensuremath{\mathrm{dev_{0}}}}
\newcommand{\devone}{\ensuremath{\mathrm{dev_{1}}}}

\newcommand{\numpassages}{\ensuremath{n}}  
\newcommand{\budget}{\ensuremath{b}}  

We first give an overview of \acrshort{ODQA} and the relevant work in adaptive computation. 

\subsection{Open Domain Question Answering}
In \acrshort{ODQA} we are given a natural language query $\query$ and a large number of passages $\corpus$---for example, all paragraphs in Wikipedia. The goal is to use $\corpus$ to produce the answer \result{}.
In extractive \acrshort{ODQA} this answer corresponds to a span in one of the documents of $\corpus$.
The corpus \corpus{} can be very large, and a common approach to reduce computational costs is to first determine a smaller document set $\docset_{\query} \subseteq \corpus$ by retrieving the most relevant $\numpassages{}$ passages using an information retrieval module. Then we run a neural reader model on this subset.
In most works, the reader model extracts answers by applying a per-passage reader to each input passage $\x_{1}, \ldots, \x_{\numpassages{}} \in \docset_{\query}$ and then apply some form of aggregation function on the per-passage answers to produce a final answer.  
Note that the passage reader can either produce an answer span as output, or 
$\noanswer$ in case the passage does not contain an answer for the given question. 

\subsection{Transformers for ODQA}
Most current \acrshort{ODQA} models rely on transformer-based architectures~\citep{DBLP:conf/nips/VaswaniSPUJGKP17}, usually pre-trained, to implement the \passagereader{} passage reader interface.
In such models, an input passage is processed via a sequence of transformer layers; in the following, we denote the $i$-th transformer layer in the sequence as $\layer_i$.
Let $\hiddent_i$ be the input to the $i$-th transformer layer and $\hiddent_{i+1} = \layer_i(\hiddent_i)$ its output. We set $\hiddent_1 = \x$ to be the input passage.
In standard non-adaptive Transformer-based models, we incrementally build a \emph{tower}---a composition of Transformer layers---until we reach some pre-defined height $n$ and use an output layer to produces the final output,  $\result=\outputlayer(\hiddent_n)$.
In this work, due to efficiency reasons, we restrict ourselves to pre-trained ALBERT~\citep{DBLP:conf/iclr/LanCGGSS20} models. One critical property of these models is parameter tying across layers: $\layer_i(\hiddent)=\layer_j(\hiddent)$ for any $i,j$.

\subsection{Adaptive Computation}
Our goal is to early-exit the iterative layer-by-layer process in order to save computation.
We assume this can be happening adaptively, based on the input, since some passages might require less computation to produce an answer than others.
\citet{DBLP:conf/acl/SchwartzSSDS20} show how this can be achieved for classification tasks. They first require internal layers to be able to produce outputs too, yielding an \emph{anytime} algorithm.~\footnote{In practice, \citet{DBLP:conf/acl/SchwartzSSDS20} choose a subset of layers to be candidate output layers, so strictly speaking we cannot exit any time, but only when a candidate layer is reached.} This can be achieved with a suitable training objective. Next, for each candidate layer $i$, they calculate the 
exit probability given its hidden state $\hiddent_i$, and use them for taking an early-exit decision: if the highest exit probability is above a global threshold $\threshold$, they return $\outputlayer(\hiddent_i)$ otherwise they continue with the following layers.
The output layer probabilities 
are not 
calibrated for exit decisions, and hence \citet{DBLP:conf/acl/SchwartzSSDS20} tune them on an held-out validation set via temperature calibration~\citep{DBLP:conf/icml/GuoPSW17,DBLP:journals/corr/abs-2003-07892}, where a temperature $T$ is tuned to adapt the softmax output probabilities at each layer. 

\section{Adaptive Computation in ODQA} \label{sec:adaptive}

Our goal is to incrementally build up towers of transformer layers for all passages in $\docset_{\query}$ in a way that minimises unnecessary computation. Our algorithms maintain a state, or \emph{skyline}, $\state=(\curheight, \currepr)$, consisting of current tower heights $\curheight=(\height_1, \ldots, \height_n)$, indicating how many layers have been processed for each of the $n$ towers, and the last representations $\currepr = (\hidden_1, \ldots, \hidden_n)$ computed for each of the towers.
We want to build up the skyline so that we reach an accurate solution fast and then stop processing.

\subsection{Early Exit with Local Exit Probabilities} \label{ssec:local}
Our first proposal is to extend the method from \citet{DBLP:conf/acl/SchwartzSSDS20} in order to build up the skyline \state{}. In particular, we will process each passage $\x_{i} \in \docset_{\query}$ in isolation, building up height $\height_i$ and representation $\hidden_i$ until an \emph{exit probability} reaches a threshold. For \citet{DBLP:conf/acl/SchwartzSSDS20} the exit probability is set to be the probability of the most likely class. While \acrshort{ODQA} is not a classification problem per se, it requires solving one as a sub-step, either explicitly or implicitly: 
deciding whether a passage contains the answer. 
In turn, our first method \localmethod{}, uses the probability $1 - \hasanswer(\hidden_i)$ of the passage not containing the answer to calculate the exit probability at such given layer. In practice the probability $\hasanswer(\hidden_i)$ is calculated as the \sigmoid{} output of an MLP applied the representation of the \CLS{} token in $\hidden_i$. Moreover, models are trained to produce $\hasanswer$ probabilities for each layer using a per-layer loss.
Following \citet{DBLP:conf/acl/SchwartzSSDS20}, we also conduct temperature calibration for the $\hasanswer$ modules using the development set. 

When building up the towers, \localmethod{} produces early exit decisions for each tower in isolation. 
Once all towers have been processed, the method selects the highest $\numoutputs$ towers in the final $\state^*$ to produce the final answer, where $\numoutputs$ is a hyperparameter.
Since some of the selected towers in $\state^*$ may not have full height, we will need to continue unrolling them to full height to produce an answer. We will call this the $\lastlayer$ strategy. Alternatively, we can return the solution at the current height, provided that we use an anytime model not just for $\hasanswer$ predictions but also for answer extraction. We will refer to this strategy as $\alllayer$. 
By default we use \lastlayer{} but we will conduct ablation study of these two approaches in \cref{sec:alllayer_vs_lastlayer}.

\subsection{Global Scheduling}\label{ssec:global}
We can apply \localmethod{} independently to each passage $\x_{i} \in \docset_{\query}$. 
However, if we have already found an answer after building up one tower for a passage $\x_{i}$, we can avoid reading other passages.
Generally, we imagine that towers that are more likely to produce the answers should be processed first and get more layers allocated to.
To assess if one tower is more likely to contain an answer, we need to compare them and decide which tower has highest \emph{priority}.
This type of strategy cannot be followed when processing passages in isolation, and hence we consider a global multi-passage view.
A simple approach for operating on multiple passages is to re-use information provided to the \localmethod{} method and select the next tower to extend using the $\hasanswer$ probabilities. In particular, we can choose the next tower to build up as $j = \argmax_i \hasanswer(\hidden_i)$, and then set  $\hidden_j \leftarrow \layer(\hidden_j)$ and $h_j \leftarrow h_j + 1$ in the state $\state$. To efficiently implement this strategy 
we use a priority queue. Every time a tower is expanded, its $\hasanswer$ probability is re-calculated and used in a priority queue we choose the next tower from. 
Once we reach the limit of our computation budget, we can stop the reading process and return the results of the highest \numoutputs{} towers $\state^*$ as inputs to its \outputphase{} phase.
The two aforementioned answer extraction methods (\ie{} \alllayer{} and \lastlayer{}) also apply to this method.

\subsection{Learning a Global Scheduler} \label{ssec:learning} 
Using \hasanswer{} probabilities to prioritise towers is a sensible first step, but not necessarily optimal. 
First, while the probabilities are calibrated, they are tuned for optimising the negative log-likelihood, not the actual performance of the method. Second, the \hasanswer{} probability might not capture everything we need to know about the towers in order to make decisions. For example, it might be important to understand what the rank of the tower's passage is in the retrieval result, as higher ranked passages might be more fruitful to expand. 
Finally, the \hasanswer{} probabilities are not learnt with the global competition of priorities across all towers, so they are not optimal for comparing priorities between towers that have different heights.

To overcome the above issues, we frame the tower selection process as a reinforcement learning (RL) problem: we consider each tower $i \in \{1,\ldots,n\}$ as a candidate action, and learn a policy $\pi(i|\state)$ that determines which tower to expand next based on the current skyline. We present the corresponding details below. 

\subsubsection{Policy}
Our policy calculates $\pi(i | \state)$ using a priority vector $\mathbf{\priority}(\state)\in \mathbb{R}^n$. The priority $\priority_i(\state)$ of each tower $i$ is calculated using a linear combination of the $\hasanswer$ probability of that tower and the output of a multi-layer perceptron $\mlp_{\theta}$. The perceptron is parametrised by $\theta$ and uses a feature representation $\mathbf{f}_i(\state)$ of tower $i$ in state $\state$ as input. Concretely, we have:
\begin{equation*} \label{eq:priority}
    \priority_i(\state) = \alpha \hasanswer(\hidden_i) + \mlp_{\theta}(\mathbf{f}_i(\state))
\end{equation*}
\noindent where $\alpha$ is a learnable mixture weight. 
As feature representation we use $\mathbf{f}_i(\state)=\left[\heightembed{}(\height{}_i), \rankembed{}(i), \hasanswer(\hidden_i) \right]$ where the tower height $\height{}_i$ and index $i$ are represented using embedding matrices $\heightembed \in \mathbb{R}^{l \times d} $ and $\rankembed \in \mathbb{R}^{n \times d}$ respectively.
When a tower is currently empty, an initial priority $\priority_i^0$ will be provided: it can either be a fixed value or a learnable parameter, and its impact is analysed in \cref{ssec:glocal}.
Given the above priority vector, the policy simply maps per tower priorities to the probability simplex:
\begin{equation*}
\pi(i|\state) = \softmax_{i}(\mathbf{\priority}(\state)).
\end{equation*}
The parameters $(\alpha, \theta)$ introduced by this policy do not introduce much computational overhead: with embedding size $d = 8$ and using $32$-dimensional hidden representations in the MLP, this model only introduces 1,039 new parameters, a small amount compared to ALBERT ($\approx$ 18M).
\subsubsection{Training}
While executing a policy, the scheduler needs to make discrete decisions as which tower to pursue. These discrete decisions mean we cannot simply frame learning as optimising a differentiable loss function. Instead we use the REINFORCE algorithm~\citep{Williams:92} for training our policy, by 
maximising the expected cumulative reward. 
For us, this reward is defined as follows. Let $\mathbf{i}_1^m=i_1,\ldots, i_m$ and $\mathbf{\state}_1^m=\state_1,\ldots,\state_m$ be a trajectory of (tower selection) actions and states, respectively. We then set the cumulative reward to $R(\mathbf{i}_t^m,\mathbf{\state}_t^m)=r(i_t,\state_t) + \gamma R(\mathbf{i}_{t+1}^m,\mathbf{\state}_{t+1}^m)$ where $r(i_t,\state_t)$ is a immediate per-step reward we describe below, and $\gamma$ is a discounting factor. 

We define an immediate per-step reward $r(i,\state)$ of choosing tower $i$ in state $\state$ as $r(i,\state)=r-c$ where $r=1$ if the selected tower contains an answer and $r=0$ otherwise.  $c \in \mathbb{R}_{+}$ is a penalty cost of taking a step. In our experiments, we set $c = 0.1$.  

\section{Related Work}

\paragraph{Adaptive Computation}
One strategy to reduce a model's complexity 
consists in dynamically deciding which layers to execute during inference~\citep{DBLP:journals/corr/BengioBPP15,DBLP:journals/corr/Graves16}. 
\emph{Universal transformers}~\citep{DBLP:conf/iclr/DehghaniGVUK19} 
can learn after how many layers to emit an output conditioned on the input.
\citet{DBLP:conf/iclr/ElbayadGGA20} generalise universal transformers by also learning which layer to execute at each step. 
\citet{DBLP:conf/acl/SchwartzSSDS20,DBLP:conf/acl/LiuZWZDJ20} propose methods that can adaptively decide when to early stop the computation in sentence classification tasks.
To the best of our knowledge, previous work has focused adaptive computation for a single input. We are the first to learn how to prioritise computation across instances 
in the context of \acrshort{ODQA}. 

\paragraph{Smaller Networks}
Another strategy 
consists in training smaller and more efficient models.
In \emph{layer-wise dropout}~\citep{DBLP:conf/emnlp/LiuRSG0018}, during training, layers are randomly removed, making the model robust to layer removal operations.
This idea was expanded \citet{DBLP:conf/iclr/FanGJ20} to modern Transformer-based models.
Other methods include \emph{Distillation}~\citep{DBLP:journals/corr/HintonVD15} of a teacher model into a student model, \emph{Pruning} of architectures after training~\citep{DBLP:conf/nips/CunDS89} and \emph{Quantisation} of the parameter space~\citep{DBLP:journals/corr/abs-1805-10796,DBLP:journals/corr/abs-1909-05840,DBLP:journals/corr/abs-1910-06188}. 
These methods are not adaptive, but could be used in concert with the methods proposed here.
\paragraph{Open Domain Question Answering}
Most modern \acrshort{ODQA} systems adopt a two-stage approach that consists of a retriever and a reader, such as DrQA~\citep{DBLP:conf/acl/ChenFWB17}, HardEM~\citep{DBLP:conf/emnlp/MinCHZ19},  BERTserini~\citep{DBLP:conf/naacl/YangXLLTXLL19}, Multi-passage BERT~\citep{DBLP:conf/emnlp/WangNMNX19}, and PathRetriever~\citep{DBLP:conf/iclr/AsaiHHSX20}.
As observed by \citet{DBLP:conf/acl/ChenFWB17,DBLP:conf/naacl/YangXLLTXLL19,DBLP:journals/corr/abs-2004-04906,DBLP:conf/emnlp/WangNMNX19}, the accuracy of such two-stage models increases with more passages retrieved.
But it remains a challenge to efficiently read a large number of passages as the reader models are usually quite computationally costly.

\section{Experiments} \label{sec:exp}

\paragraph{Dataset}
SQuAD-Open~\citep{DBLP:conf/acl/ChenFWB17} is a popular open-domain question answering dataset based on SQuAD. 
We partition the dataset into four subsets: training set, two development sets (\devzero{} and \devone{}), and test set, and their details are summarised in \cref{tab:datasize}. 

\begin{table}[t]
\resizebox{\columnwidth}{!}{
    \begin{tabular}{lcccc}
      \toprule
      {\bf SQuAD-Open} & { train} \ & { \devzero{} } \ & { \devone{} } \ & { test} \\
      \midrule
      Size & 78,839 & 4,379 & 4,379 & 10,570  \\
      Hits@30 & 71.2\% & 72.7\% & 72.1\% & 77.9\% \\
      \bottomrule
    \end{tabular}
}
\caption{Dataset sizes and retriever performances.} \label{tab:datasize}
\end{table}

\paragraph{Experimental Setup}

We follow the preprocessing approached proposed by \citet{DBLP:conf/emnlp/WangNMNX19} and split passages into 100-word long chunks with 50-word long strides. We use a BM25 retriever 
to retrieve the top $\numpassages$ passages for each question as inputs to the reader and the Wikipedia dump provided by \citet{DBLP:conf/acl/ChenFWB17} as source corpus.  Following \citet{DBLP:conf/emnlp/WangNMNX19}, we set $\numpassages=5$ for training and  $\numpassages = 30$ for test evaluations.
\cref{tab:datasize} shows the Hits@30 
results of our BM25 retriever on the dataset and they are comparable with previous works~\citep{DBLP:conf/naacl/YangXLLTXLL19,DBLP:conf/emnlp/WangNMNX19}.

\paragraph{Reader Model}
For all our experiments, we fine-tune a pre-trained ALBERT model~\citep{DBLP:conf/iclr/LanCGGSS20}, consisting of 24 transformer layers and cross-layer parameter sharing.
We do \emph{not} use global normalisation~\citep{DBLP:conf/acl/GardnerC18} in our implementation, but our full system (without adaptive computation) achieves an EM score of 52.6 and is comparable to Multi-passage BERT~\citep{DBLP:conf/emnlp/WangNMNX19} which uses global normalisation.

\paragraph{Training Pipeline}
The anytime reader models are first trained on training set and validated on \devzero{}. Then we conduct temperature calibration on \devzero{}. 
For \globalmethodRL{}, the scheduler model is trained on \devzero{} with the calibrated anytime model, and validated with \devone{}.

\paragraph{Baselines}
Following \citet{DBLP:conf/acl/SchwartzSSDS20}, we use three types of baselines: 
\begin{inparaenum}[\itshape 1\upshape)]
    \item the \emph{standard baseline} that reads all passages and outputs predictions at the final layer,
    \item the \emph{efficient baseline} that always exits at a given intermediate layer for all passages, and is optimised to do so, 
    \item the \emph{top-$k$ baseline} that only reads the $k$ top ranked passages and predicts the answer at their final layers.
\end{inparaenum}

\begin{figure*}[t]
\begin{center}
\begin{subfigure}{.49\textwidth}
\includegraphics[width=\textwidth]{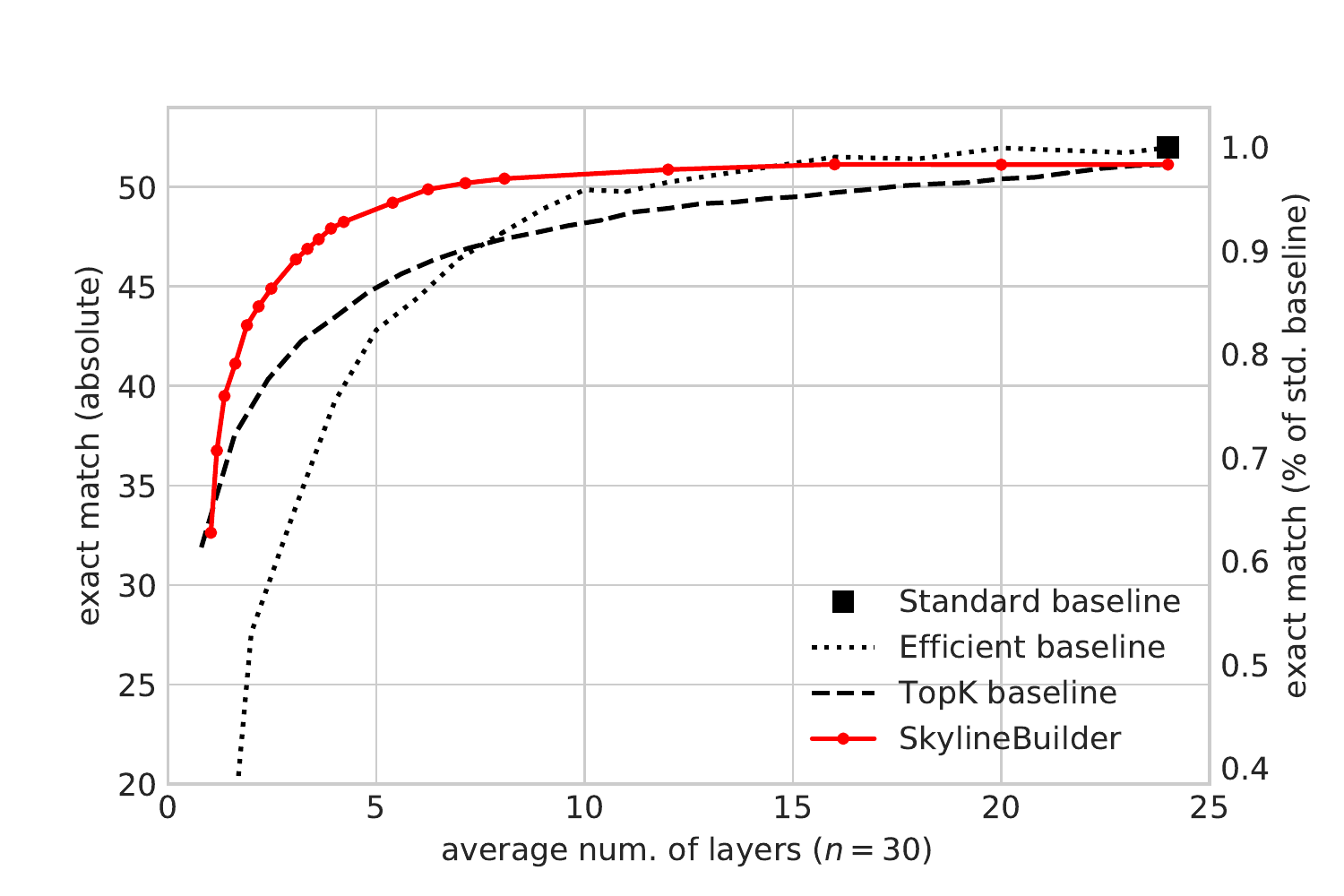}
\subcaption{\globalmethodRL{} vs. baselines} \label{fig:master}
\end{subfigure}
\begin{subfigure}{.49\textwidth}
\includegraphics[width=\textwidth]{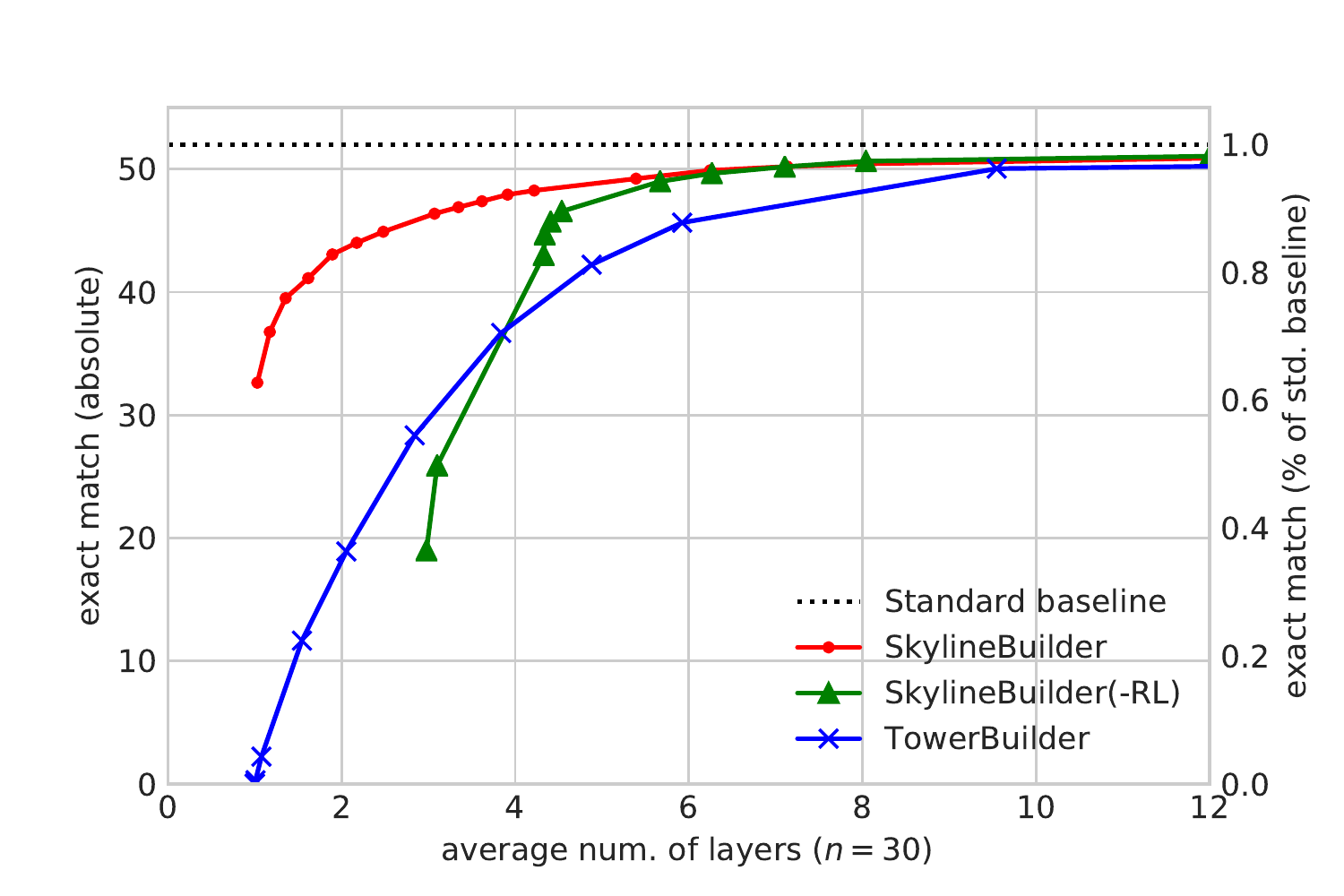}
\subcaption{Local vs. Global Models} \label{fig:local_vs_global}
\end{subfigure}
\begin{subfigure}{.49\textwidth}
\includegraphics[width=\textwidth]{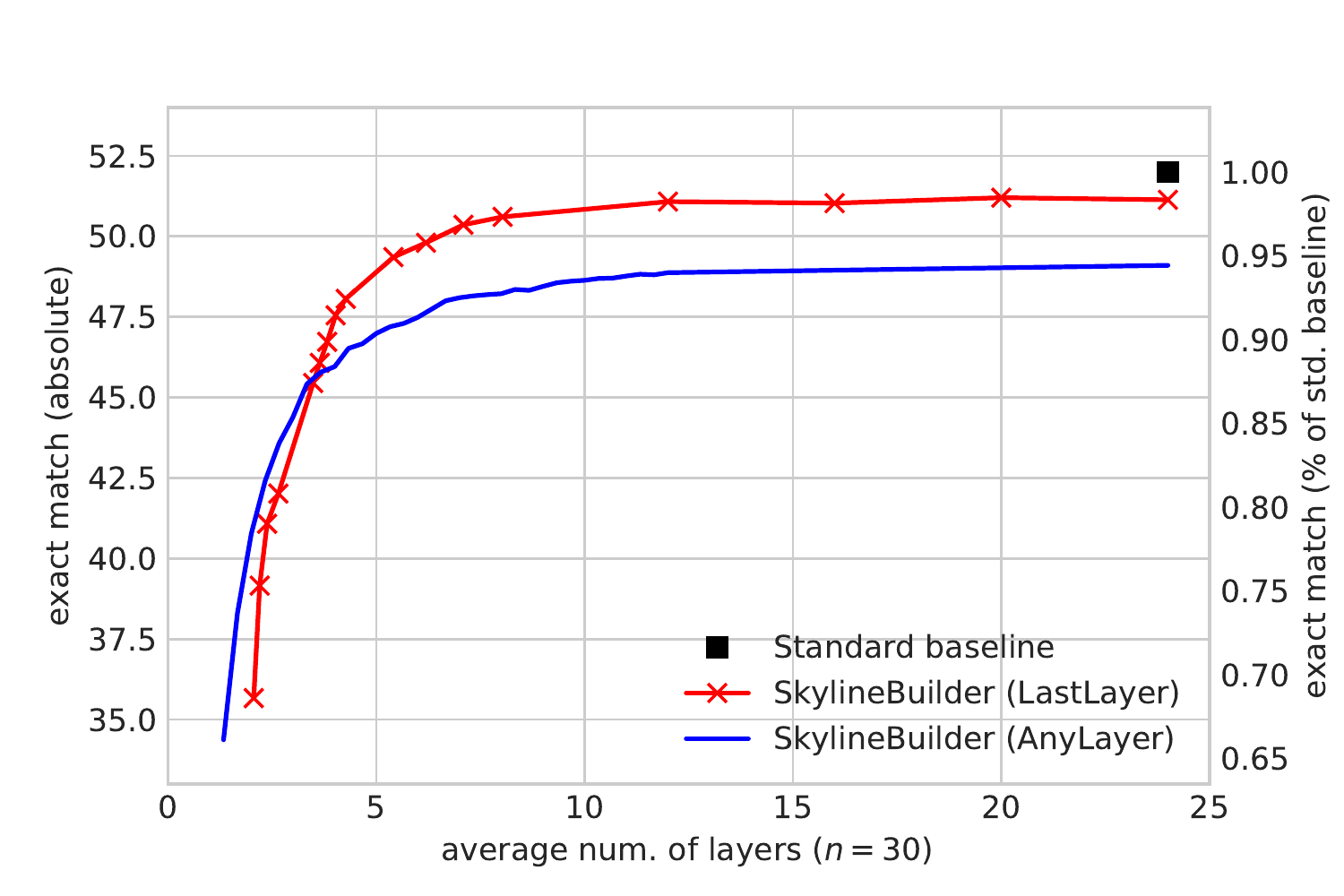}
\subcaption{\alllayer{} vs. \lastlayer{}} \label{fig:anytime_vs_lastlayer}
\end{subfigure}
\begin{subfigure}{.49\textwidth}
\includegraphics[width=\textwidth]{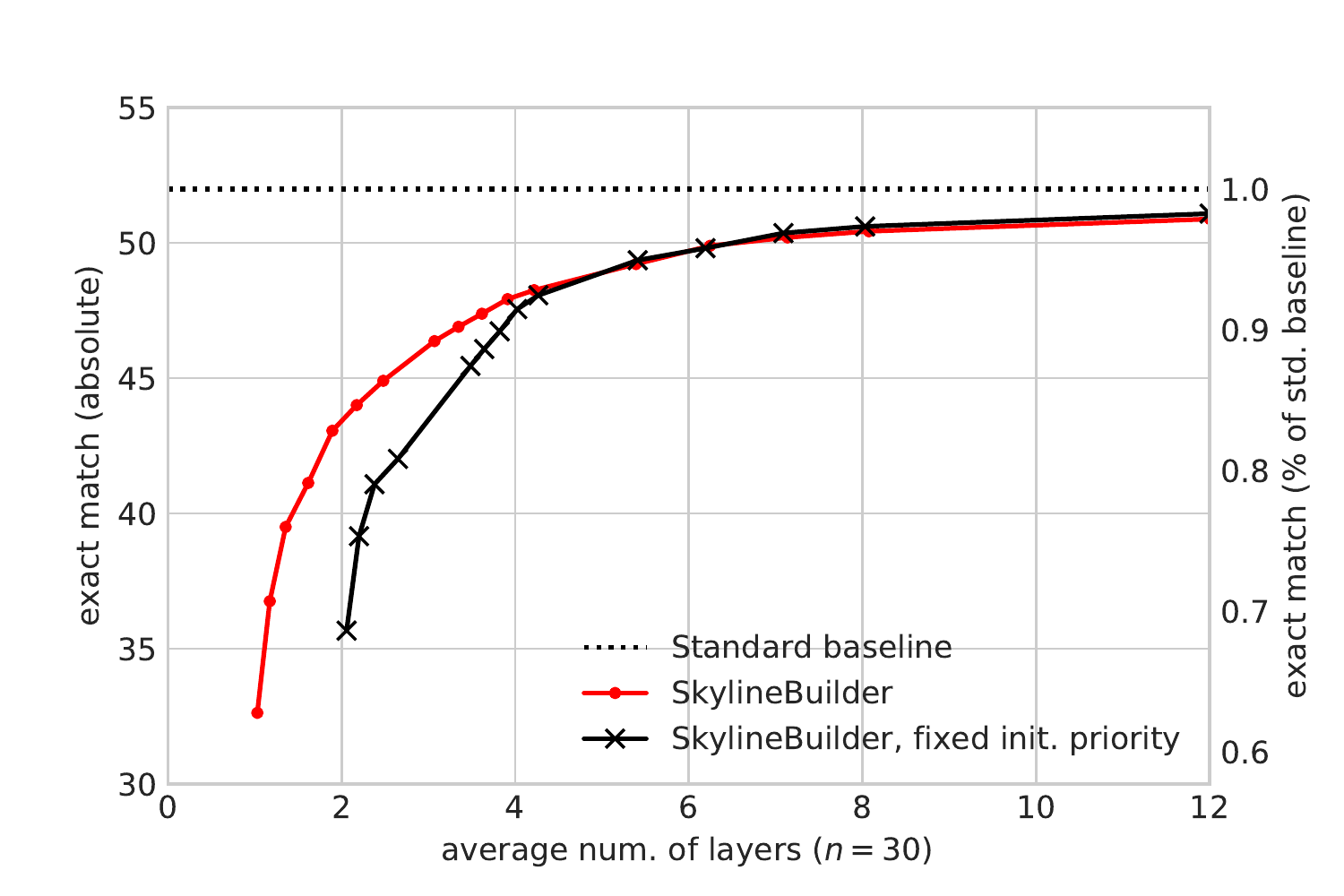}
\subcaption{Learnt vs. Fixed Initial Priorities} \label{fig:init_priority}
\end{subfigure}
\end{center}
\caption{Evaluation results on the SQuAD-Open test set with 30 passages.} \label{fig:results}
\end{figure*}

\paragraph{Evaluation protocol}
Our goal is to assess the computational efficiency of a given method in terms of accuracy vs. computational budget used. 
We follow \citet{DBLP:conf/iclr/FanGJ20} and consider the computation of one layer as a unit of computational cost. In particular, we will assess how many layers, on average, each method builds up for each passage. 
Similarly to \citet{DBLP:conf/acl/SchwartzSSDS20}, we show the accuracy-efficiency trade-off for different strategies by showing the computation cost on the $x$-axis, and the Exact Match (EM)~\footnote{The evaluation script can be found at this address: \href{https://github.com/facebookresearch/DrQA}{https://github.com/facebookresearch/DrQA}.} score on the $y$-axis.
\subsection{Static vs. Adaptive Computation}
We first investigate how adaptive computation compares to the static baselines. We will focus on a single adaptive method, \globalmethodRL{}, and assess different adaptive variants later. 

\cref{fig:master} shows the accuracy of \globalmethodRL{} at different budgets when compared to the standard, efficient, and top-$k$ baselines. 
We note that it reaches the similar results of the static baselines with much fewer layers.
In particular, it yields substantially higher performance than static methods when the computational budget is smaller than ten layers. For example, when given four layers on average, \globalmethodRL{} achieves EM score 48.0, significantly outperforming EM score 44.2 of the top-$k$ baseline.
In \cref{tab:budget_em} we consider a setting where \globalmethodRL{} and the static baseline reach comparable (95\%) performance of the full 24-layer model.
We see that simply reducing the number of passages to process is giving a poor accuracy-efficiency trade-off, requiring 14.4 layers (or 18 passages) to achieve this accuracy. 
The efficient baseline fares better with 9.5 layers, but it is still outperformed by \globalmethodRL{}, that only needs 5.6 layers on average to reach the desired accuracy.

\begin{table}[t]
\resizebox{\columnwidth}{!}{
    \begin{tabular}{lcc}
      \toprule
      {\bf Method} & {\bf Avg. \#layers} \ & {\bf Reduction } \  \\
      \midrule
      Standard baseline & 24 & 1.0x  \\
      Efficient baseline & 9.5 & 2.5x  \\
      Top-$k$ baseline & 14.4 & 1.7x  \\
      \midrule
      \localmethod{} & 9.0 & 2.7x  \\
      \globalmethod{} & 6.1 & 3.9x  \\
      \globalmethodRL{} & \bf{5.6} & \bf{4.3x}  \\
      \bottomrule
    \end{tabular}
}
\caption{Reduction in layer computations while achieving 95\% of the accuracy of the standard baseline.} \label{tab:budget_em}
\end{table}

\begin{table*}[!hbt]
\begin{center}
\resizebox{\textwidth}{!}{
\begin{tabular}{lcccccc}
  \toprule
   & { $\text{Var}(h)$} \ & { $\text{Avg}(\text{rank})$ } \ & { Flips} \ & { $h_{+}-h_{-}$} \ & { HAP} \ & { Exact Match} \\
  \midrule
  Efficient Baselines & 0.00 & 14.50 & - & 0.00 & 6.1\% & 23.47 \\
  \midrule
  \localmethod{} & 11.05 & 13.38 & - & 3.68 & 22.0\% & 17.10 \\
  \midrule
  \globalmethod{} & 7.46 & 13.06 & 13.37 & 3.46 & 27.4\% & 27.95 \\
  \globalmethodRL{} & 12.71 & 8.78 & 6.48 & 5.99 & {40.5}\% & {\bf 33.60} \\
  \bottomrule
\end{tabular}
}
\caption{Quantitative analysis on SQuAD Open \devone{} set with top 30 passages and two layers of computation per passage on average.} \label{tab:analysis}
\end{center}
\end{table*}

\subsection{Local vs. Global Models} \label{ssec:glocal}
What is the impact of globally selecting which towers to extend, rather than taking early-exit decisions on a per-tower basis? 
To answer this question, we consider two global methods: \globalmethodRL{} and \globalmethod{}, 
the method in \cref{ssec:global} that uses $\hasanswer$ probabilities as priorities without any RL-based selection policy.
We compare both to the local method \localmethod{}. 

\cref{fig:local_vs_global} shows that, while for very low budgets \localmethod{} outperforms \globalmethod{}, with a budget larger than 4 layers it is not the case anymore.
This may be due to a tendency of \globalmethod{} spending an initial computation budget on exploring many towers---in \cref{fig:tower_vis} we show examples of this behaviour.
It is also shown that \globalmethodRL{} considerably outperforms both \localmethod{} and \globalmethod{}.
Along with the results in \cref{tab:budget_em}, the comparisons above indicate that 
\begin{inparaenum}[\itshape 1\upshape)]
    \item global scheduling across multiple towers is crucial for improving efficiency, and
    \item optimising the adaptive policy with RL manage to exploit global features for tower selection, leading to further improvements.
\end{inparaenum}

\subsection{Ablation Studies}
\paragraph{Any Layer vs. Last Layer Model} \label{sec:alllayer_vs_lastlayer}
For comparing the $\lastlayer$ and the $\alllayer$ strategies introduced in \cref{ssec:local}, we show the behaviour of these methods for the \globalmethodRL{} scheduling algorithm in \cref{fig:anytime_vs_lastlayer}.
Using an anytime answer extraction model has a negative effect on accuracy. We see this clearly at 24 layers where $\alllayer$ lags substantially behind the standard baseline while $\lastlayer$  almost reaches it. 
We see this gap across the whole budget spectrum, leading to less accurate results except for very small budgets.
\paragraph{Learning Initial Priorities}
\globalmethodRL{} uses a learnt initial priority for each tower. This not only enables it learn which towers to process first at the beginning, but also how long to wait until other towers are visited. 
\cref{fig:init_priority} shows the benefit gained from adopting this strategy: without training the initialisation priorities, \globalmethodRL{} spend more computation on passages that are likely not needed.
Once an average of 4 layers have been added, the benefit disappears as \globalmethodRL{} with learnt initial priorities will try to visit more candidates itself.  
\subsection{Quantitative Analysis}
This section aims at understanding where and how our adaptive strategies behave differently, and what contributes to the gain in the accuracy-efficiency trade-off.
We propose the following quantitative metrics:
\begin{inparaenum}[\itshape 1\upshape)]
    \item $\text{Var}(h)$: variance of the heights of the towers. 
    \item $\text{Avg}(\text{rank})$: average rank of the tower when the method chooses which tower to build on.
    \item Flips: how often does the strategy switch between towers, measuring the exploration-exploitation trade-off of a method. 
    \item $h_{+}-h_{-}$: $h_{+}$ (resp. $h_{-}$) is the average height of towers with (resp. without) an answer. Their difference measures the difference in amount of computation between passages with the answer and the ones without an answer.
    \item \hasanswer{} Precision (HAP): how often a tower selection action selects a tower whose passage contains the answer.
\end{inparaenum}
We analyse our proposed methods along with static baselines on the SQuAD development set; results are outlined in \cref{tab:analysis}.
Overall, the higher the $\hasanswer$ Precision, the more accurate the method.
This finding matches with our intuition that, if a tower selection strategy can focus its computation on passages that contain the answer, it yields more accurate results with smaller computation budgets. 

Comparing \globalmethod{} and \globalmethodRL{} gives more insights regarding what the RL training scheme learns.
\globalmethodRL{} learns a policy with the highest $\text{Var}(h)$, the lowest $\text{Avg}(\text{rank})$, and the lowest number of tower flips, suggesting that 
\begin{inparaenum}[\itshape 1\upshape)]
    \item it focuses on a few towers rather than distributing its computation over all passages,
    \item it is more likely to select top-ranked passages, and 
    \item it switches less between towers, and tends to build one tower before switching to another.
\end{inparaenum}
\globalmethodRL{} also yields the highest $\hasanswer$ Precision and $h_{+}-h_{-}$, meaning that tends to prioritise the passages containing the answer. 

\begin{figure}[!tb]
\begin{center}
\begin{subfigure}{\columnwidth}
\begin{center}
\fbox{\includesvg[height=3.2cm]{figures/visualise/towers/skylinebuilder_39.svg}}
\fbox{\includesvg[height=3.2cm]{figures/visualise/towers/skylinebuilderRL_39.svg}}
\end{center}
\subcaption{Example 1} \label{ex:1}
\end{subfigure}
\begin{subfigure}{\columnwidth}
\begin{center}
\fbox{\includesvg[height=3.2cm]{figures/visualise/towers/skylinebuilder_1231.svg}}
\fbox{\includesvg[height=3.2cm]{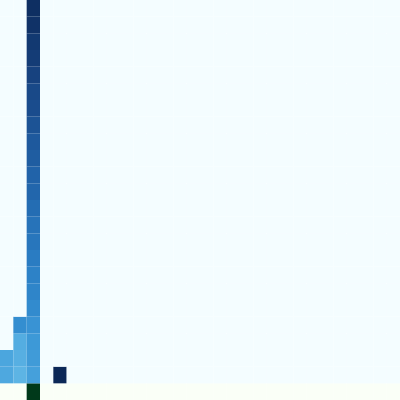}}
\end{center}
\subcaption{Example 2} \label{ex:2}
\end{subfigure}
\begin{subfigure}{\columnwidth}
\begin{center}
\fbox{\includesvg[height=3.2cm]{figures/visualise/towers/skylinebuilder_117.svg}}
\fbox{\includesvg[height=3.2cm]{figures/visualise/towers/skylinebuilderRL_117.svg}}
\end{center}
\subcaption{Example 3} \label{ex:3}
\end{subfigure}
\end{center}
\caption{Examples of the skylines built by \globalmethod{} (left) and \globalmethodRL{} (right), with two layers per passage on average. 
The green blocks indicate towers that contain the answer. 
} \label{fig:tower_vis}
\end{figure}

\subsection{Qualitative Analysis and Visualisation}
Here we analyse how different methods build the skyline.
\cref{fig:tower_vis} shows some examples of skylines built by \globalmethod{} and \globalmethodRL{}. 
The towers are ordered by the rank of their associated passages in the retrieval results from left to right, and are built bottom-up. 
The colour gradient of the blues blocks reflects the order in which the layers are built: darker cells correspond to layers created later in the process. 

In \cref{ex:1} and \cref{ex:2} we can see that \globalmethodRL{} tends to focus on one or two towers,
whereas \globalmethod{} 
has a more even distribution of computation across different towers.
In \cref{ex:2}, even when only one tower contains the answer, \globalmethodRL{} manages to locate it and build a full-height tower on it.

\cref{ex:3} shows a case where none of the top 4 passages contains the answer. \globalmethodRL{} goes over these irrelevant towers quickly and start exploring later towers, until it reaches the tower with rank 27 and becomes confident enough to keep building on it.
These examples shows how \globalmethodRL{} learns an efficient scheduling algorithm to locate passages containing the answer with very limited budgets.

\begin{figure}[t]
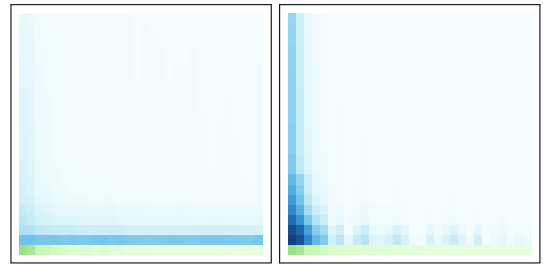

\begin{center}
\begin{subfigure}{\columnwidth}
\begin{center}
\fbox{\includesvg[height=3.2cm]{figures/visualise/heatmaps/skylinebuilder_heatmap.svg}}
\fbox{\includesvg[height=3.2cm]{figures/visualise/heatmaps/skylinebuilderRL_heatmap.svg}}
\end{center}
\end{subfigure}
\end{center}
\caption{Heatmap of the tower selections by \globalmethod{} (left) and \globalmethodRL{} (right). The colour gradient of the blues blocks reflects their selection frequencies.
} \label{fig:heatmap}
\end{figure}

To understand how our proposed methods work at macro level, we use heat-maps (\cref{fig:heatmap}) for showing how frequently each block is selected.
The green row at the bottom indicates the frequency of each passage containing the answer.
\globalmethod{} explores all passages quite evenly, whereas \globalmethodRL{} learns to prioritise top-ranked towers. This preference is reasonable because, as shown by the green row at the bottom, top-ranked towers are more likely to contain the answer. 
Also note that \globalmethodRL{} does not naively process towers from left to right like the top-$k$ baseline does, but instead it learns a trade-off between \emph{exploration and exploitation}, leading to the significant improvement over the top-$k$ baseline shown in \cref{fig:master}.

\subsection{Adaptive Computation vs. Distillation} \label{sec:compare_distil}

Distillation is another orthogonal approach to reduce computational cost. We compare our adaptive computation method \globalmethodRL{} with a static DistilBERT~\citep{DBLP:journals/corr/abs-1910-01108} baseline, and the results are shown in \cref{tab:distilbert}. Our method significantly outperforms DistilBERT while computing much fewer layers. 

\begin{table}[t]
\resizebox{\columnwidth}{!}{
    \begin{tabular}{lcc}
      \toprule
      {\bf Models} & {\bf Num. layers} & {\bf EM} \\
      \midrule
      DistilBERT~\citep{DBLP:journals/corr/abs-1910-01108} & 6 & 40.5 \\
      \globalmethodRL{} & 1.6 & 41.1 \\
      \globalmethodRL{} & 3 & 46.4 \\
      \globalmethodRL{} & 6 & {\bf 49.7} \\
      \bottomrule
    \end{tabular}
}
\caption{Comparing adaptive computation with distillation on SQuAD-Open test set.} \label{tab:distilbert}
\end{table}

\section{Discussions and Future Works}
In this paper, we focus on reducing the number of layers and operations of \acrshort{ODQA} models, but the actual latency improvement also depends on the hardware specifications.
On GPUs we cannot expect a reduction in the number of operations to translate 1:1 to lower execution times, since they are highly optimised for parallelism.~\footnote{When evaluated on an NVIDIA TITAN X GPU, our proposed \globalmethodRL{} achieves approximately 2.6x latency reduction while retaining 95\% of the performance.}
We leave the parallelism enhancements of \globalmethodRL{} for future work.

We also notice that the distillation technique is complementary to the adaptive computation methods. It will be interesting to integrate these two approaches to achieve further computation reduction for \acrshort{ODQA} models.

\section{Conclusions}
In this work we show that adaptive computation can lead to substantial efficiency improvements for \acrshort{ODQA}. In particular, we find that it is important to allocate budget dynamically across a large number of passages and prioritise different passages according to various features such as the probability that the passage has an answer. Our best results emerge when we learn prioritisation policies using reinforcement learning that can switch between exploration and exploitation. On our benchmark, our method achieves 95\% of the accuracy of a 24-layer model while only needing 5.6 layers on average.

\subsubsection*{Acknowledgements}
This research was supported by the European Union's Horizon 2020 research and innovation programme under grant agreement no. 875160.

\bibliography{bibliography}
\bibliographystyle{acl_natbib}

\clearpage

\appendix

\section{Experimental Details} \label{sec:appendix_experiments}

\subsection{Hyper-parameters}

\begin{table}[!hb]
\begin{center}
    \begin{tabular}{lc}
      \toprule
      {\bf Hyper-parameter} & {\bf Value} \\
      \midrule
      learning rate & 3e-5 \\
      weight decay & 0.01 \\
      batch size & 48 \\
      epoch & 2 \\
      optimiser & Adam \\
      Adam $\epsilon$ & 1e-6 \\
      Adam $(\beta_1, \beta_2)$ & (0.9, 0.999) \\
      warmup ratio & 10\% \\
      max sequence length & 200 \\
      max question length & 100 \\
      max answer length & 30 \\
      number of passages & 5 \\
      dropout & 0.0 \\
      pretrained model & \href{https://huggingface.co/albert-large-v2}{albert-large-v2} \\
      number of parameters & 18M \\
      device & Nvidia Titan X \\
      \bottomrule
    \end{tabular}
\caption{Hyper-parameters for reader model training.} \label{tab:hyperparam}
\end{center}
\end{table}

\begin{table}[!hb]
\begin{center}
    \begin{tabular}{lc}
      \toprule
      {\bf Hyper-parameter} & {\bf Value} \\
      \midrule
      learning rate & 1e-3 \\
      batch size & 32 \\
      epoch & 16 \\
      optimiser & SGD \\
      max number of steps & 240 \\
      step cost $c$ & 0.1 \\
      discount factor $\gamma$ & 0.9 \\
      number of passages & 30 \\
      \bottomrule
    \end{tabular}
\caption{Hyper-parameters for scheduler model RL training.} \label{tab:hyperparamRL}
\end{center}
\end{table}

\end{document}